\RequirePackage{fix-cm}
\documentclass[twocolumn]{svjour3}       
\smartqed  
\usepackage{graphicx}
\usepackage{subfigure}
\journalname{Journal XXX}
\begin{document}

\title{Modeling reverse thinking for machine learning}


\author{Huihui Li\and Guihua Wen* 
}


\institute{H. Li \and G. Wen (Corresponding author)\at
             School of Computer Science and Engineering, South China University of Technology, China \\
              Tel.: +86-18998384808\\
              \email{crghwen@scut.edu.cn}           
}

\date{}

\maketitle

\begin{abstract}

Human inertial thinking schemes can be formed through learning, which are then applied to quickly solve similar problems later. However, when problems are significantly different, inertial thinking generally presents the solutions that are definitely imperfect. In such cases, people will apply creative thinking, such as reverse thinking, to solve problems. Similarly, machine learning methods also form inertial thinking schemes through learning the knowledge from a large amount of data. However, when the testing data are vastly difference, the formed inertial thinking schemes will inevitably generate errors. This kind of inertial thinking is called illusion inertial thinking. Because all machine learning methods do not consider illusion inertial thinking, in this paper we propose a new method that uses reverse thinking to correct illusion inertial thinking, which increases the generalization ability of machine learning methods. Experimental results on benchmark datasets are used to validate the proposed method.

\end{abstract}

\section{Introduction}
Human learning automatically induces more general rules from a large number of experiences. These rules are taken as the elements to form the inertial thinking schemes that will be used to solve similar problems later. The reasons that inertial thinking exists for humans lie in the fact that it can solve a large number of daily problems when there is insufficient information and it can solve them quickly \cite{Corcoran2009}. However, the accuracy of solutions to the problems will generally be reduced when the problems are different from those for learning.

Like human learning, machine learning automatically learns new concepts from a large number of data. In the learning process, inertial thinking schemes will be formed, including illusion inertial thinking, leading to low generalization ability of the machine learning methods. According to the law of Hoeffding \cite{Hoeffding1963}, the accuracy of machine learning methods depend heavily on a large number of training samples, and this is why most advanced machine learning methods require a large number of training samples \cite{Lake2015}, e.g., deep learning methods \cite{LeCun2015}. However, it is difficult to obtain a large number of training samples for most real applications, such as disease diagnosis, where the training samples are few, easily leading to formation of illusion inertial thinking in machine learning methods. One way for humans to solve these kinds of problem is the use of reverse thinking, which is an effective method of creative thinking \cite{Sawaguchi2015}. However, for machine learning methods, reverse thinking is totally not considered. Therefore, in this paper we present a new method that applies reverse thinking to correct illusion inertial thinking for machine learning methods so that their generalization ability can be enhanced. This method is suitable for any machine learning method, as long as illusion inertial thinking is facilitated.
\subsection{Machine learning principles}
Machine learning methods are all based on the principle of inertial thinking. They first learn from the training data to form the inertial thinking scheme that is then applied to solve new problems.

Suppose that for the testing sample $X$, a learning approach is applied to obtain the probability model $P$ from the training database, so that
\begin{equation}
C^*=\arg\max_{C_i\in C}P(C_i=X),
\end{equation}
where $C=\{C_1,\cdots,C_i,\cdots,C_n\}$ is a set of classes and $P(C_i=X)$ is the probability that $X$ belongs to class $C_i$.
For example, if there are two classes  $C_1,C_2$, the learned probability model will work as follows:
\begin{displaymath}
P(C_1=X)> P(C_2=X)\Longrightarrow X\in C_1,
\end{displaymath}
\begin{displaymath}
P(C_2=X)>P(C_1=X)\Longrightarrow X\in C_2
\end{displaymath}

The ability of the learned probability model to correctly classify the new samples is measured as its generalization ability. A lower generalization ability of the model may result from the training samples being insufficient to correctly fit the data distribution of the problem.
In order to formally illustrate the generalization ability of the model, all data are taken as balls of different colors in a jar. It is expected that the proportion of different color balls in the jar can be computed through a portion of the balls partially taken out of the jar. It assumes that the ratio of orange balls in the jar is $\mu$ and the proportion of green balls for $1-\mu$. In the balls taken out of the jar, the ratio of orange balls is $\nu$, and the proportion of green balls is $1-\nu$. The relationship between the two cases follows the Hoeffding inequality \cite{Hoeffding1963}:
\begin{equation}
P(|\mu-\nu|>\varepsilon)\leq 2exp(-2\varepsilon^2N).
\end{equation}

This law shows that when \emph{N} is large, the right-hand side of the inequality will be very small, indicating that the probability of the difference between $\mu$ and $\nu$ being larger than the given $\varepsilon$ is very small. In such a case, the distribution of the overall samples can be inferred from the distribution of the partially taken samples. However, in real applications, \emph{N} is usually difficult to be sufficiently large, so error is inevitable.
The current machine learning methods only consider suitable inertial thinking, ignoring illusion inertial thinking. In such a case, when the size of the training samples are not large enough, the testing samples could have a high probability of differing from the training samples, inevitably leading to error. This is why there are many machine learning methods that perform well on some experimental data, with even up to 100\% accuracy, but they may behave badly in practice.

\section{Modeling reverse thinking}
\begin{figure*}[htbp]
 \centering
  \includegraphics[width=\textwidth]{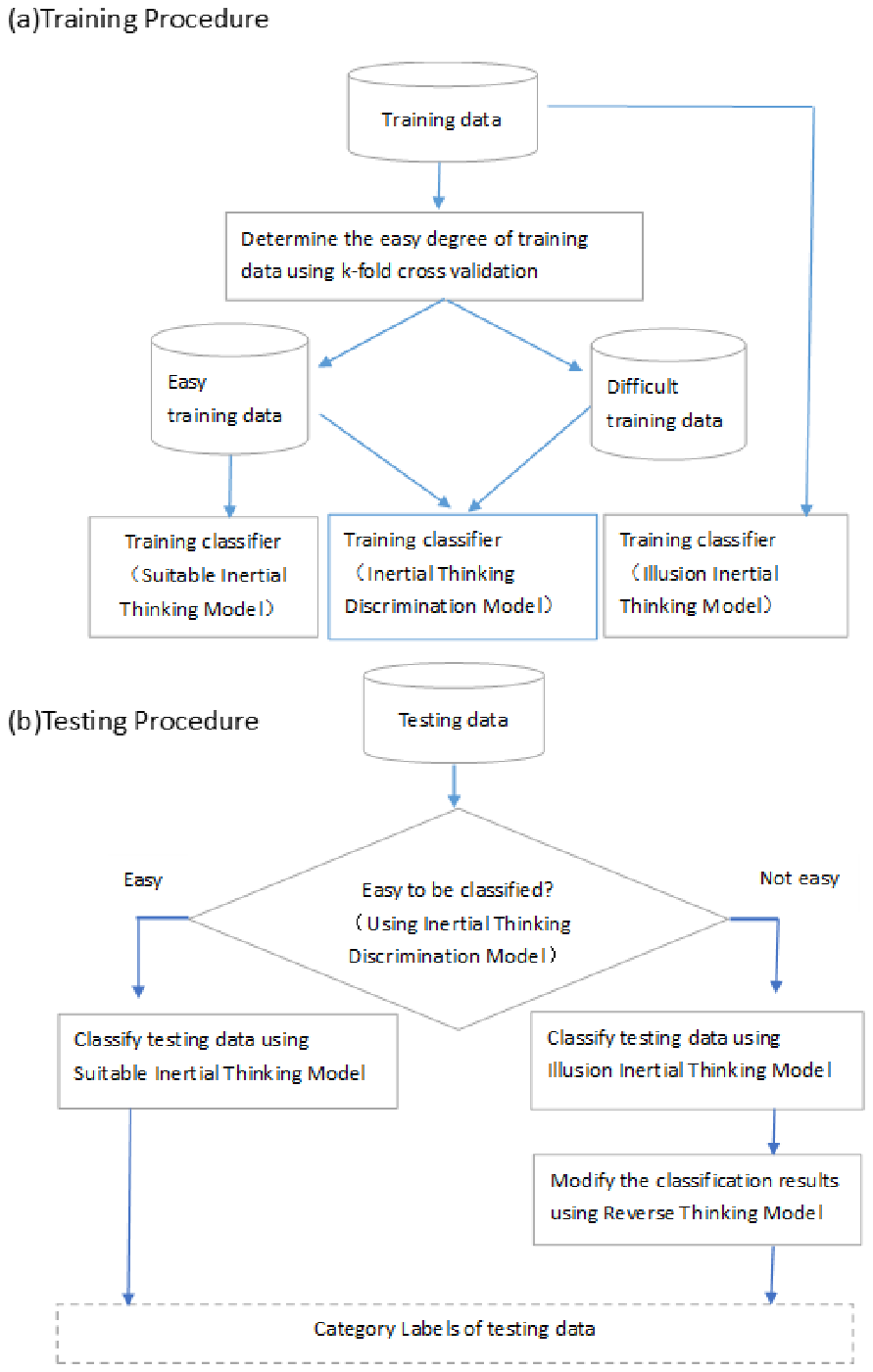}
\caption{Training and testing procedures of machine learning methods based on reverse thinking}
\label{Figure1}
\end{figure*}

Generally, inertial thinking is composed of both suitable inertial thinking and illusion inertial thinking. When the illusion inertial thinking appears, people can use reverse thinking to deal with it. This process begins with designing the inertial thinking discrimination model to judge whether the current testing data is in line with inertial thinking or illusion inertial thinking. Second, the reverse thinking model is created to correct illusion inertial thinking. The overall procedure is shown in Fig. \ref{Figure1}, where classifiers are trained on the training database to form a suitable inertial thinking model, an illusion inertial thinking model, and an inertial thinking discrimination model.

\subsection{Inertial thinking model}
The suitable inertial thinking model works well on the simple samples that are easily classified, while illusion inertial thinking easily occurs when classifying complicated samples. There are many measures that can be applied to rank samples by simplicity \cite{Smith2014}. Hence, we rank samples in the training data according to the simplicity of samples using the k-fold cross-validation method. When samples in a fold are removed from the training data, the remaining ones are used to train the classifier, which is then applied to classify all training data. In such a case, the simplicity of each sample can be computed.

\textbf{Definition}. Supposing that there are $N$ classification models, the simplicity of sample $X$ is defined by

\begin{equation}
    m(X)=\frac{1}{N}\sum_{i=1}^N m_i(X)
\end{equation}
\begin{displaymath}
    m_i(X)=  \left\{
               \begin{array}{ll}
                 1, & \hbox{X is classified correctly;} \\
                 0, & \hbox{else.}
               \end{array}
             \right.
\end{displaymath}

The definition of simplicity means that a sample is simple if it is correctly classified by many models. In this way, all samples can be ranked according to their simplicity:
\begin{displaymath}
    S=\{X_1,...,X_i,...,X_n|m(X_i)\geq m(X_{i+1})\}.
\end{displaymath}

Subsequently, a simple training database $(S_E,L_E)$ can be constructed, the samples of which are of greater simplicity than the given threshold $\theta$. Similarly, a complicated training database $(S_D,L_D)$ can also be constructed:

\begin{displaymath}
    S_E=\{X_i|Xi\in S \wedge m(X_i)> \theta\},
\end{displaymath}
 \begin{displaymath}
    S_D=\{X_i|Xi\in S \wedge m(X_i)\leq \theta\},
\end{displaymath}
\begin{displaymath}
    S=S_E\cup S_D, S_E\cap S_D=\emptyset.
\end{displaymath}

Using these training databases, three models are trained as follows, where $\xi$ is a classification method and $\varphi$ is a discrimination method for judging whether a sample is simple.
\begin{itemize}
\item Inertial thinking model: $M_I=\xi(S,L)$.
\item Suitable inertial thinking model: $M_S=\xi(S_E,L_E)$.
\item Discrimination model: $M_{D}=\varphi(S_E\cup S_D,\{+,-\})$ , where $+$ denotes samples to be classified easily and $-$ denotes the samples to be classified difficultly.
\end{itemize}

\subsection{Reverse thinking rules}
Owing to illusion inertial thinking, the learned inertial thinking model may misclassify the coming samples. For example, it classifies the test sample $X$ with true class $C_1$  as belonging to the class $C_2$; that is,
\begin{equation}
    r:C_1\rightarrow C_2.
\end{equation}

In such a case, according to reverse thinking, the above rule can be reversed to obtain the reverse thinking rule by recomputing the probability:
\begin{equation}
    rr:C_2\rightarrow C_1.
\end{equation}

Because the results obtained by illusion inertial thinking are not absolute, a probabilistic model like the Bayesian theorem is applied. In conditions of the complete statistical knowledge of a model presented, the classifier designed according to the Bayesian theorem can obtain the minimum classification error in terms of probability, which means that the classifier based on the reverse thinking rule is optimal.

As the testing sample $X$ has been misclassified as belonging to the class $C_2$, according to the Bayesian theorem its true class is $C_1$ with the probability
\begin{equation}
    P(C_1=X|C_2=X)=\frac{P(C_2=X|C_1=X)P(C_1=X)}{P(C_2=X)},
\end{equation}
where $P(C_i=X)$ refers to the probability that $X$ belongs to class $C_i$, and $P(C_2=X|C_1=X)$ denotes the probability that $X$ with the class $C_2$ is misclassified as belonging to $C_1$. The right-hand side of the above equation involves the prior probability that is then applied to compute the  posterior probability of the left-hand side of the equation. Equation (6) is applied to implement the reverse thinking rule in Eq. (5) by recomputing the probability. All posterior probabilities can be computed from the confusion matrix. To avoid confusion, Eq. (6) can be rewritten as
\begin{equation}
    P(C_1=X|C_2=X)=\frac{P(C_2|C_1)P(C_1=X)}{P(C_2=X)}.
\end{equation}

\textbf{Theorem 1}. If $X\in C_2\wedge P(C_1=X) > P(C_2=X)$, under certain conditions, $P(C_2=X|C_1=X)>P(C_1=X|C_2=X)$, so that $X\in C_1$

\textbf{Proof}. The inertial thinking model classifies the sample with the true class $C_1$ as belonging to the class $C_2$, and the reverse thinking rule can be applied to recompute the probability so as to classify the sample as belonging to the correct class $C_1$; that is,
\begin{displaymath}
    P(C_1=X|C_2=X)=\frac{P(C_2|C_1)P(C_1=X)}{P(C_2=X)}
\end{displaymath}
However, the reverse thinking rule may be also applied to recompute the probability
\begin{displaymath}
    P(C_2=X|C_1=X)=\frac{P(C_1|C_2)P(C_2=X)}{P(C_1=X)}.
\end{displaymath}

In such a case,
\begin{displaymath}
    \frac{P(C_1=X|C_2=X)}{P(C_2=X|C_1=X)}=
\end{displaymath}
\begin{displaymath}
    \frac{P(C_2|C_1)P(C_1=X)}{P(C_2=X)}\times \frac{P(C_1=X)}{P(C_1|C_2)P(C_2=X)}
\end{displaymath}
\begin{displaymath}
   =\frac{P(C_2|C_1)}{P(C_1|C_2)}\times (\frac{P(C_1=X)}{P(C_2=X)})^2.
\end{displaymath}

To satisfy $X\in C_1$,
\begin{displaymath}
   \frac{P(C_2|C_1)}{P(C_1|C_2)}\times (\frac{P(C_1=X)}{P(C_2=X)})^2 >1,
\end{displaymath}
\begin{equation}
   \frac{P(C_2|C_1)}{P(C_1|C_2)} > (\frac{P(C_2=X)}{P(C_1=X)})^2,
\end{equation}

\begin{displaymath}
   \frac{P(C_2|C_1)}{P(C_2=X)} > \frac{P(C_1|C_2)}{P(C_1=X)}\times \frac{P(C_2=X)}{P(C_1=X)}.
\end{displaymath}

The learning model has classified $X$ as belonging to $C_2$ -- that is, $P(C_2=X)>P(C_1=X)$ -- leading to
\begin{displaymath}
   \frac{P(C_2|C_1)}{P(C_2=X)} > \frac{P(C_1|C_2)}{P(C_1=X)}.
\end{displaymath}

Thus, the reverse thinking rule can enhance the probability that the classifier will classify $X$ as belonging to $C_1$ more than it will classify it as belonging to $C_2$. This proves that the reverse thinking rule has the ability to correct the classification of samples that have been misclassified.

Since $P(C_2=X)>P(C_1=X)$, their difference is less, the inequality (8) is easily satisfied, and then the reverse thinking rule is more useful in correcting the misclassified category. However,
\begin{displaymath}
   \frac{P(C_2|C_1)}{P(C_1|C_2)} > 1,
\end{displaymath}
which means that the sample with class $C_1$ is easily misclassified as belonging to the class $C_2$ in the confusion matrix, but not \textit{vice versa}.

As the classifier classified the samples depending on $P(C_1)=X$ and $P(C_2=X)$, they may be wrong, so they can be also replaced by the prior probability computed in the confusion matrix. Subsequently, we have another alternative reverse thinking rule, as
\begin{equation}
  P(C_1=X|C_2=X)=\frac{P(C_2|C_1)P(C_1)}{P(C_2)}.
\end{equation}

It is noteworthy that the reverse thinking rule is only applied to the classification of complicated samples, in which only one reverse direction is considered to deal with the misclassified samples. Therefore, in the following context, the necessary condition for the reverse thinking rule is simplified as follows to design the new machine learning approach:
\begin{equation}
    P(C_1=X|C_2=X)>P(C_1=X)\Rightarrow X\in C_1,
\end{equation}
indicating that the reverse thinking rule can enhance the probability that the classifier correctly classifies the sample.

\subsection{Computing prior probabilities}
It is necessary to use the Bayesian theorem to calculate the probabilities of reverse thinking rules, which can be calculated by computing the confusion matrix of the specified classifier on the given training data. For example, if the training data have 60 samples with two categories, where each category is composed of 30 samples, the confusion matrix of the given classifier can be computed through 10-fold cross-validation. As shown in Table \ref{Table1}, each column in the table represents the predicted category, and the total number of each column represents the number of samples predicted for the category. Each row represents the true category of the sample, and the total number of samples per row represents the number of samples in that class. For example, the number, 23, located at the cross-point of both the first row and first column indicates that the samples with the true class $C_1$ are correctly predicted to be $C_1$. The number, 5, located at the cross-point between the first column and second row indicates that there are five samples with true class $C_2$, which are incorrectly predicted to be class $C_1$.

\begin{table}[htbp]
\centering
\caption{Confusion matrix}
\label{Table1}
\begin{tabular}{llll}
\hline\noalign{\smallskip}
 Predicted class          &         &      &        \\
        $\Downarrow$      &         &$C_1$ & $C_2$  \\
\noalign{\smallskip}\hline\noalign{\smallskip}
                          & $C_1$	& 23	& 7 \\
True class$\Rightarrow$   & $C_2$	&5 	    &25\\
\noalign{\smallskip}\hline
\end{tabular}
\end{table}

According to the confusion matrix, the prior conditional probabilities that the samples are misclassified as belonging to each category can be computed by
\begin{equation}
    P(C_1=X|C_2=X)=\frac{23}{7+25},
\end{equation}
\begin{equation}
    P(C_2=X|C_1=X)=\frac{25}{23+5}.
\end{equation}

\subsection{RTML algorithm}
The reverse thinking for machine learning (RTML) algorithm is described in Table \ref{Table2}, including both training and testing stages. The inertial thinking model in the RTML algorithm is obtained by training the classifier on all the training data. This is because illusion inertial thinking is formed in the original environment. Second, for any new testing sample, the discrimination model is used to decide whether the testing sample is easy to classify. If the testing sample is easy to classify, it is classified by using the suitable inertial thinking model that is trained on the easy training data. If the testing sample is hard to classify, the inertial thinking model will be applied, and then the reverse thinking model is applied to modify its result. The final category is determined by the modified results.

\begin{table}[htbp]
\centering
\caption{Reverse thinking for machine learning (RTML) algorithm}
\label{Table2}
\begin{tabular}{rlll}
\hline\noalign{\smallskip}
Input &training data $(S,L)$ and testing sample $X$\\
      &The threshold parameter $\theta$\\
Output& The class label $\omega$ for X\\
\noalign{\smallskip}\hline\noalign{\smallskip}
Training &\\

1 &Compute $m(X_i),X_i\in S$ \\
2 &Build $(S_E,L_E ) \wedge X_i\in S_E$ $\wedge m(X_i)<\theta$\\
3 &Build $(S_D,L_D )\wedge  X_i\in S $$\wedge m(X_i)\geq\theta$\\
4 &Train suitable inertial thinking model \\
  &$M_S=\xi(S_E,L_E)$ \\
5 &Train illusion inertial thinking model\\
  &$M_I=\xi(S,L)$\\
6 &Train the discrimination model\\
  & $M_{II}=\varphi(S_E\cup S_D,\{+,-\})$\\
\noalign{\smallskip}\hline\noalign{\smallskip}
Testing&\\
7&$y\leftarrow M_{II}(X)$\\
8&If $y\in \{+\},\omega\leftarrow M_S(X)$\\
9& Else /* reverse thinking is required*/\\
10 &$\omega\leftarrow M_{I}(X)$\\
11 & If $\omega = C_1$ and \\
   & $P(C_2=X|C_1=X) > P(C_2=X)$\\
12 & $\omega\leftarrow C_2$\\
13 & End\\
14 & If $\omega = C_2$ and\\
   & $P(C_1=X|C_2=X) > P(C_1=X)$\\
15 & $\omega\leftarrow C_1$\\
16 & End\\
\noalign{\smallskip}\hline
\end{tabular}
\end{table}

In this algorithm, only two classes are considered because most of the datasets in real tasks are binary class. Second, for binary-class datasets, the use of reverse thinking rules only needs to consider the Bayesian theorem, without considering the Bayesian network, so as to simplify the problem. Finally, the multi-class tasks can be classified through classification methods on binary-class tasks, such as the support vector machine (SVM) \cite{Ji2017}.

The complexity of the RTML algorithm at the training stage is mainly distributed in computing the confusion matrix and training three classifiers. Therefore, the complexity of the RTML algorithm is equal to the maximum complexity of the three classifiers. The complexity of the RTML algorithm at the testing stage is also the maximum complexity of the three classifiers for testing a sample. The calculation of reverse thinking rules is only related to the number of categories, which can be regarded as a constant. The RTML algorithm has an additional parameter, the simplicity threshold, which determines the boundary between the illusion inertial thinking model and suitable inertial thinking model.

\section{Experimental results}

The effectiveness of the proposed method is demonstrated through experiments on standard datasets. In principle, the proposed method is applicable to all existing classification methods, so as to improve their classification performance. In experiments, a highly representative machine learning method is selected, namely SOFTMAX, because it is widely used in deep learning \cite{Szegedy2014,LeCun2015}, and it is expected that the RTML algorithm can be applied to deep learning. For the discrimination model, SVM-KNN (where KNN denotes \textit{k}th nearest neighbor) is selected \cite{Zhang2006}, as it inherits the SVM advantages \cite{Ji2017}, while be able to elegantly deal with large datasets.

\subsection{Experimental databases}
Experimental databases include artificial datasets and benchmark real datasets.

\begin{figure}[htbp]
  \centering
  \subfigure[SOFTMAX]{
    \includegraphics[width=2.0 in]{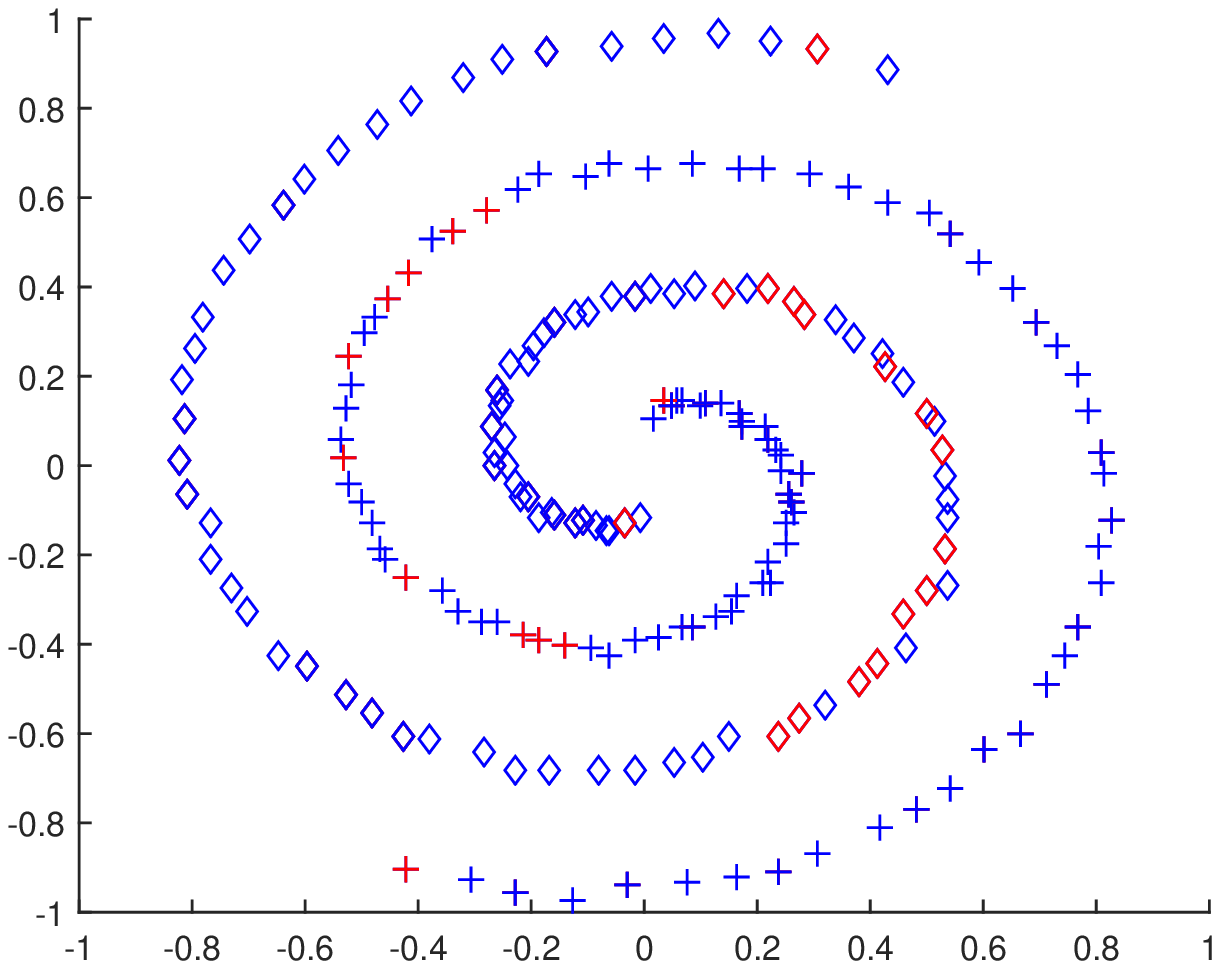}}
  \hspace{0 in}
  \subfigure[RTML]{
    \includegraphics[width=2.0 in]{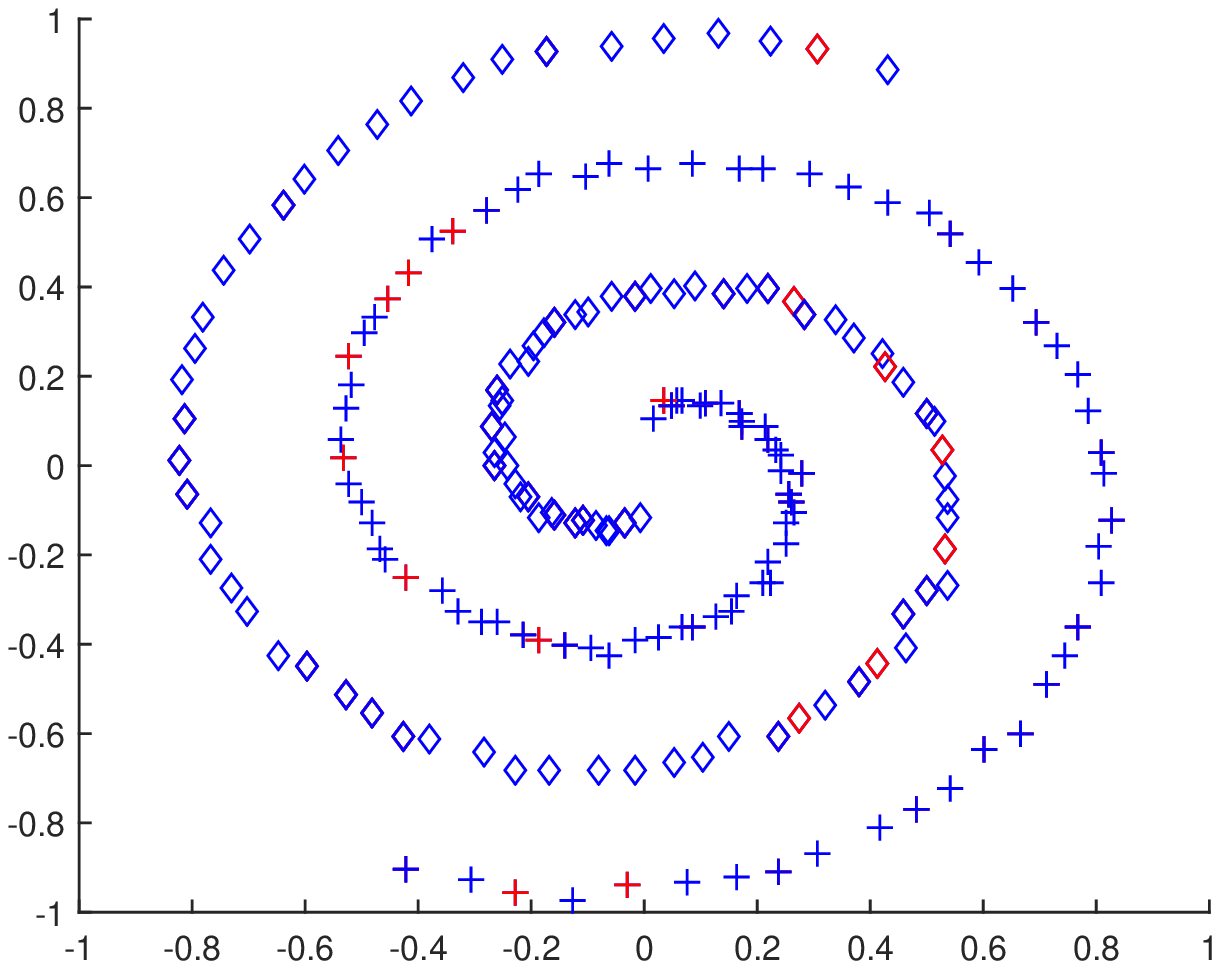}}
  \caption{Classification results in which misclassified samples are indicated in red}
  \label{Figure2}
\end{figure}

Using artificial data, we can control the number of available samples and add noise according to the experimental purpose. In the experiments, two spirals of data are selected, which is synthetic, but very difficult for many machine learning methods. Such data have been widely used by many classifiers as standard data \cite{Wen2013}. Here, they are applied to observe the effect of reverse thinking on noisy data of different sizes. One of the data examples is shown in Fig. \ref{Figure2}, where there are 200 points with some noise disruption. Obviously, these data are not linearly separable and are nearly evenly distributed for two classes.

In addition to artificial data, 15 benchmark real datasets with two classes are selected from LIBSVM \cite{Chang2011} and KEEL \cite{Alcal-Fdez2011}; see Table \ref{Table3}. The reason for selecting data with only two classes is that most of the real datasets are binary class. Second, for binary-class datasets, the use of reverse thinking rules only needs to consider the Bayesian theorem, without considering the Bayesian network, so as to simplify the problem. Finally, the multi-class data can be classified through classification methods on binary-class data, such as the SVM \cite{Ji2017}.

\begin{table}[htbp]
\centering
\caption{Illustration of benchmark data used for experiments}
\label{Table3}
\begin{tabular}{lllll}

\hline\noalign{\smallskip}
No.	&data name	&classes	&features	&samples\\

\noalign{\smallskip}\hline\noalign{\smallskip}
1& fourclass  &2	&2	&862\\
2&	waveform  &2	&21	&5000\\
3&	svmguide1 &2	&4	&3089\\
4&	madelon	  &2	&500	&2000\\
5&	splice	  &2	&60	&1000\\
6&	thyroid   &2	&5	&215\\
7&	titanic	  &2	&3	&2201\\
8&	segment0 &2	&19	&2308\\
9&	vehicle1  &2	&18	&846\\
10&	glass1    &2	&9	&214\\
11&	pima     &2	&8	&768\\
12&	haberman  &2 &3 &306\\
13&	spambase  &2	&57	&4597\\
14&	mammographic   &2	&5	&830 \\
15&	phoneme   &2	&5	&5404\\
\noalign{\smallskip}\hline
\end{tabular}
\end{table}

\subsection{Analysis of RTML algorithm}
In order to observe the effectiveness of the RTML algorithm more clearly, experiments on two spirals of data are conducted, the results of which are presented in Fig. \ref{Figure2}, where the misclassified testing data are denoted in red. It can be seen that the number of red points misclassified by SOFTMAX is larger than that misclassified by the RTML algorithm. Some samples misclassified by SOFTMAX have been corrected by the RTML algorithm, showing that the reverse thinking model is effective. However, when the easy points are misclassified, the RTML algorithm cannot modify them. This depends on the ability of the original classifier, i.e., SOFTMAX. This, in turn, means that the original classifier should be selected to be the best as possible. Some misclassified samples cannot be modified, since in such cases illusion inertial thinking has not formed.

\begin{figure}
  \centering
  \subfigure[titanic]{
    \includegraphics[width=2.0 in]{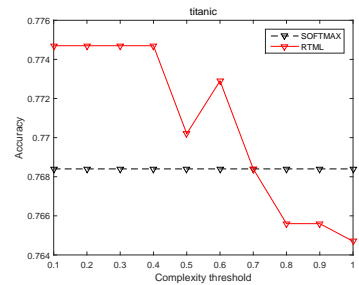}}
  \hspace{0 in}
  \subfigure[RTML]{
    \includegraphics[width=2.0 in]{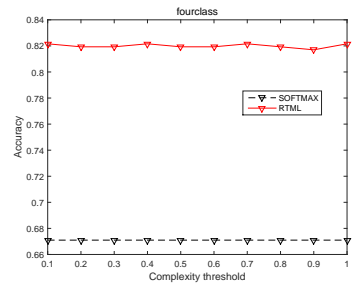}}
  \caption{Impact of the simplicity threshold $\theta$ on the performance of RTML algorithm}
  \label{Figure3}
\end{figure}

The RTML algorithm has a simplicity threshold as a parameter to decide whether the illusion inertia thinking model can be formed on the database composed of samples with a simplicity larger than the given threshold. Experiments are conducted on two real datasets to illustrate the influence of the parameter on the RTML algorithm. It can be seen from Fig. \ref{Figure3} that the RTML algorithm is sensitive to the threshold. When the threshold $\theta>0.4$ in Fig. \ref{Figure3}(a), the RTML algorithm has the worse effect, since in such a case many samples are mistaken for the easy ones. This indicates that the threshold parameter values need to be determined as accurately as possible. However, from Fig. \ref{Figure3}, the parameter varies in terms of a law, indicating that its optimal value can be determined easily using the \emph{k}-fold cross-validation method.

\begin{figure*}
    \centering
 \includegraphics[width=\textwidth]{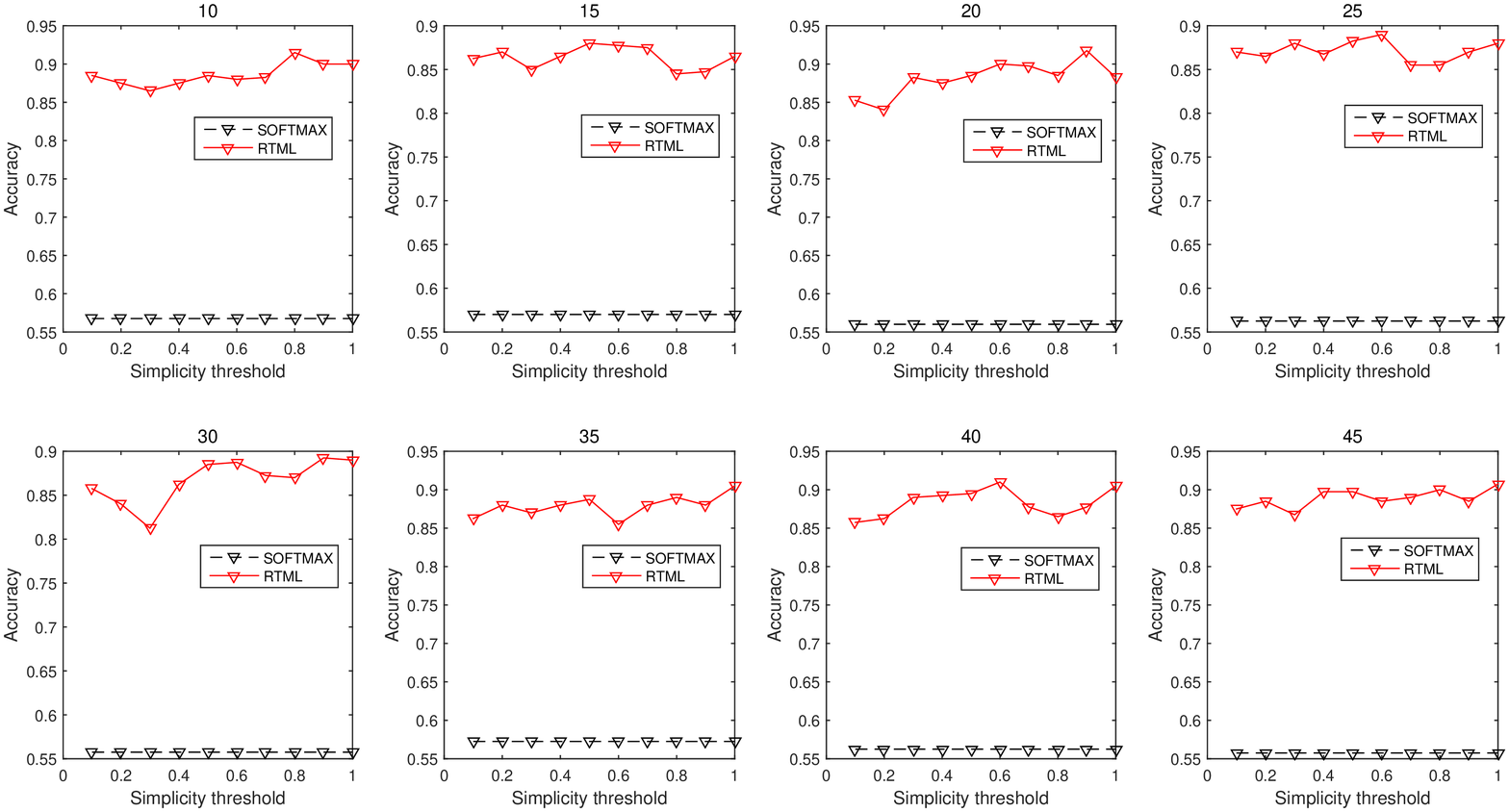}
    \caption{Experimental results of RTML algorithm with different numbers of models }
\label{Figure4}
\end{figure*}

As simplicity of samples plays a very important role in the RTML algorithm, it needs to be measured as precisely as possible. In our context, many models have been applied to define the simplicity. These models are created using the \emph{k}-fold cross-validation method on the training database, with the same classification method. It can be seen from Fig. \ref{Figure4} that when the number of models is larger than 35, the accuracy tends to the optimal one, and the threshold can be easily selected. This means that the number of models used to measure the simplicity of the samples should be large as possible to make the measurement more stable.

\subsection{Influence of noisy artificial data on RTML algorithm}

\begin{figure*}[htbp]
    \centering
 \includegraphics[width=\textwidth]{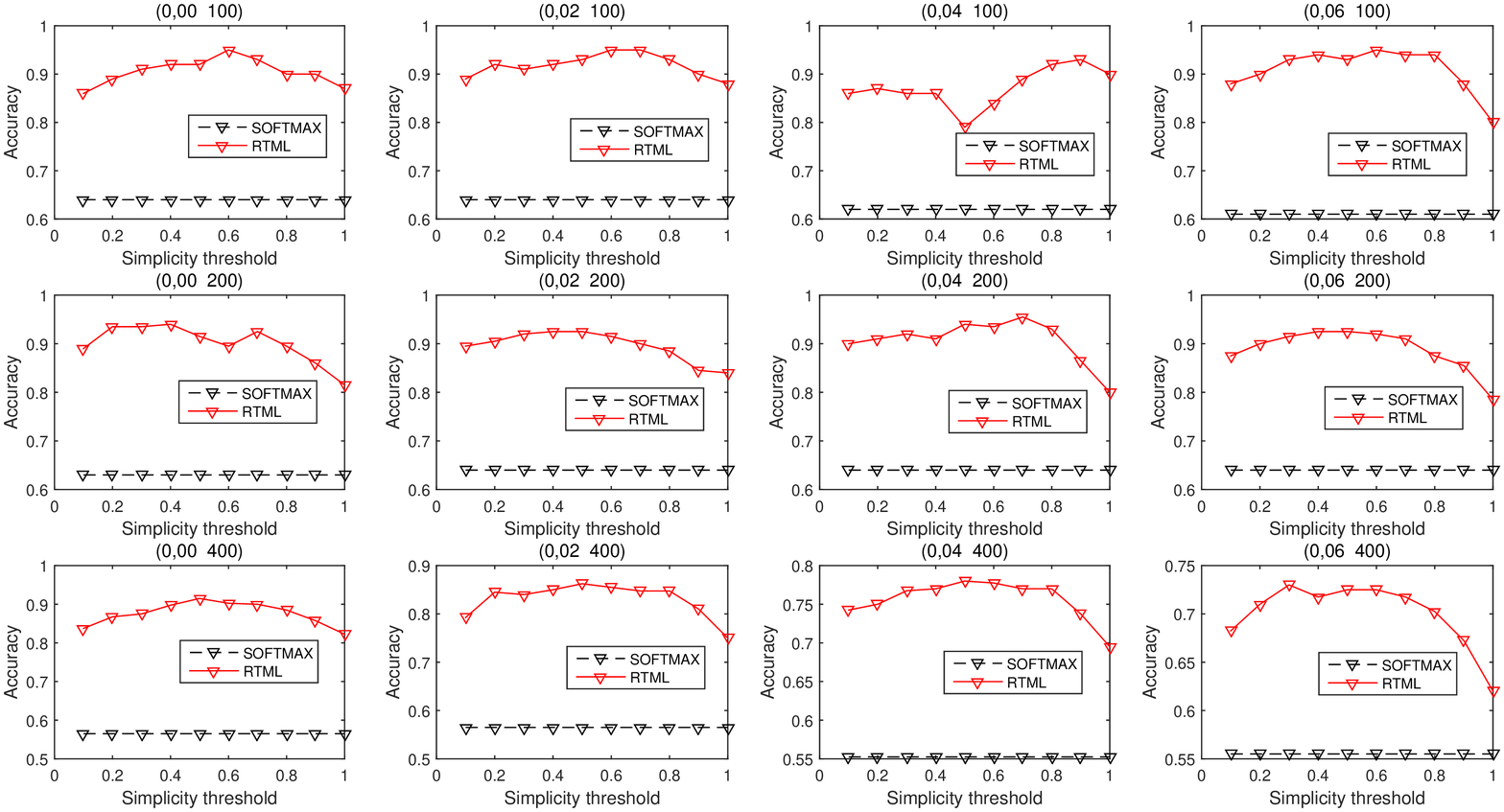}
    \caption{Experimental results of RTML algorithm with different numbers of models}
\label{Figure5}
\end{figure*}

Generally, machine learning methods are easily influenced by noisy data. Some experiments are therefore conducted to assess the influence of noisy data on the RTML algorithm, the results of which are shown in Fig. \ref{Figure5}. Here, datasets with 100, 200, and 400 points are added by noises whose variances are 0.02, 0.04, and 0.06, respectively (as well as a variance of 0.0). In this way, 12 different datasets are formed to observe the effects of the noise and the size of the data on the performance of the RTML algorithm. In order to observe the sensitivity of the RTML algorithm to its parameters at the same time, classification results corresponding to each parameter are computed and plotted in the figure. It can be seen from Fig. \ref{Figure5} that the noise has negative effects on the RTML algorithm; the more intense the noise, the lower the accuracy. This is because the noise is random, disrupting the formation of illusion inertial thinking. However, in the case of noise, the RTML algorithm still obviously increases the accuracy of SOFTMAX. At the same time, the RTML algorithm is sensitive to its parameter, but still achieves remarkable results at many values, indicating that the choice of parameter is easier. Finally, when the data size becomes larger, the RTML algorithm increases the effect more obviously while being less sensitive to its parameter. This is because the number of misclassified samples will increase in the larger dataset, leading to the formation of more stable illusion inertial thinking.

\subsection{Effect of real datasets on RTML algorithm}

As the artificially constructed examples may not correspond to situations that are likely to occur in practice, some experiments on real datasets are conducted. One of the advantages of real data is that they are generated without any knowledge of the classification procedures that will be used for testing. Here, 15 real datasets, shown in Table \ref{Table4}, are applied to conduct experiments using the 10-fold cross-validation method to determine the training data and testing data, and then the average classification accuracy can be obtained. In each set of training data, the 10-fold cross-validation method is used to select the optimal parameters for the given classifier.

\begin{table*}[htbp]
\centering
\caption{Experimental results on real datasets for which SOFTMAX is learning method}
 \label{Table4}
\begin{tabular}{lllllllllll}
 \hline\noalign{\smallskip}
    &           &SOFTMAX        &SOFTMAX       & \\
No.	&Data name	&(all data) 	&(simple data)	& RTML algorithm\\
\noalign{\smallskip}\hline\noalign{\smallskip}
1& fourclass  & 0.6893$\pm$0.0316   &0.6893$\pm$0.0316    &\textbf{0.9837$\pm$0.0103}\\
2&	waveform  &0.7742$\pm$0.0218   &0.7754$\pm$0.0231    &\textbf{0.8710$\pm$0.0236}\\
3&	svmguide1 & 0.8460$\pm$0.0140   &0.8447$\pm$0.0112   &\textbf{0.9314$\pm$0.0175}\\
4&	madelon	  & 0.5365$\pm$0.0142    &\textbf{0.5685$\pm$0.0376}  &0.488$\pm$0.0430 0\\
5&	splice	  &\textbf{0.8405$\pm$0.0235}   &\textbf{0.8405$\pm$0.0216}   &0.8368$\pm$0.0219\\
6&	thyroid   &0.8985$\pm$0.0671   &0.9212$\pm$0.0534   &\textbf{0.9396$\pm$0.0577}\\
7&	titanic	  &0.7733$\pm$0.0179&\textbf{0.7833$\pm$0.0108}	&\textbf{0.7815$\pm$0.0125}\\
8&	segment0  &0.9970$\pm$0.0036   &0.9970$\pm$0.0036   &0.9965$\pm$0.0034\\
9&	vehicle1  &\textbf{0.8062$\pm$0.0466}    &0.8003$\pm$0.0477    &0.7920$\pm$0.0573\\
10&	glass1    & 0.6397$\pm$0.0842 &0.6111$\pm$0.0516   &\textbf{0.7185$\pm$0.0522}\\
11&	pima     &0.6849$\pm$0.0355  &0.6875$\pm$0.0300 & \textbf{0.7148$\pm$0.0351}\\
12&	haberman  &\textbf{0.7448$\pm$0.0425}  &0.7414$\pm$0.0425  &0.7148$\pm$0.0787\\

13&	spambase  &  0.9180$\pm$ 0.0143   & 0.9213$\pm$0.0115 &\textbf{0.9245$\pm$0.0096}\\
14&	mammographic   & 0.7901$\pm$0.0509  & \textbf{0.7937$\pm$0.0491} &   \textbf{0.7913$\pm$0.0502}\\
15&	phoneme   &0.7541$\pm$0.0186    &0.7648$\pm$0.0174    &\textbf{0.8960$\pm$0.0154}\\
\noalign{\smallskip}\hline
     \end{tabular}
 \end{table*}

 It can be easily seen from Table \ref{Table4} that the RTML algorithm is much effective for most datasets.  The RTML algorithm's performance was especially prominent in the fourclass, svmguide1, waveform, glass1, and phoneme data, outperforming SOFTMAX in terms of accuracy by 30\%, 10\%, 10\%, 10\%, and 14\%, respectively, indicating that illusion inertial thinking definitely exists, so that the reverse thinking model can effectively correct these illusions. This can be confirmed from the confusion matrices. For example, \emph{CF(svmguide1)} is the confusion matrix of the \emph{svmguide1} data performed by SOFTMAX, where the significant illusion inertial thinking is established, as it often classifies the samples with class 1 as belonging to class 2, but not\textit{vice versa}:

$CF(svmguide1)=\left(
  \begin{array}{cc}
     0.6513  &  0.3487 \\
     0.0517 &   0.9483 \\
  \end{array}
\right)
$,

$CF(fourclass)=\left(
  \begin{array}{cc}
    0.7951  &  0.2049 \\
     0.3773  &  0.6227 \\
  \end{array}
\right)
$.

However, the RTML algorithm fails in some data, such as \emph{madelon} and \emph{splice}, performing poorly on these two data. The reason is that there the illusion inertial thinking is not obvious. It can be seen from the confusion matrix, such as \emph{CF(madelon)}, that the illusion inertial thinking schemes between the two classes are very similar. The thinking falls into chaos and cannot form stable illusion inertial thinking in one direction, leading to the failure of the reverse thinking model. In this case, good measures should be applied to maintain suitable inertial thinking and enhance illusion inertial thinking, so as to improve the overall performance. This is because strengthening the inertial thinking may also significantly strengthen the illusion inertial thinking. Both are consistent and not in conflict with each other:

$CF(splice)=\left(
  \begin{array}{cc}
            0.8354 &   0.1646\\
        0.1481  &  0.8519\\
  \end{array}
\right)
$,
$CF(madelon)=\left(
  \begin{array}{cc}
   0.7546   & 0.2454\\
    0.2516 &   0.7484\\
  \end{array}
\right)
$.

Additionally, on some datasets, such as \emph{madelon} and \emph{mammographic}, SOFTMAX performs better by taking easy samples as the training database rather than by taking all the samples as the training database. This indicates that it is reasonable to rank samples in training database by their simplicity.

\section{Conclusions}
Similar to human learning, machine learning can easily form illusion inertial thinking, but all machine learning methods do not consider it. In this paper, we use reverse thinking to overcome illusion inertial thinking, so as to improve the generalization ability of machine learning methods. The experimental results indicate that our method is very effective. The proposed method is universal, applicable to any machine learning methods, especially to those performing badly on some data, such as unbalanced data, for which machine learning methods can easily form the illusion inertial thinking. In planned future work, we will select some machine learning methods to combine with the RTML algorithm to solve concrete tasks like emotion recognition, because the training data for emotion recognition are generally unbalanced, easily inducing machine learning methods to form illusion inertial thinking. Second, the proposed method uses the Bayesian theorem, but not the Bayesian network. As a matter of fact, reverse and inertial thinking between categories are linked, and the dialectical unity between them will be investigated under the framework of the Bayesian network in the future \cite{DeMotta2016}.



\section*{Funding}
This study was supported by the China National Science Foundation (60973083/61273363), Science and Technology Planning Project of Guangdong Province (2014A010103009/2015A020217002), and Guangzhou Science and Technology Planning Project(201504291154480).

\section*{Conflicts of interest}
The authors declare that they have no conflicts of interest.

\section*{Ethical approval}
This paper does not contain any studies with human  or animals participants.


\end{document}